\documentclass[letterpaper]{article} 
\usepackage[draft]{aaai2026}  
\usepackage{times}  
\usepackage{helvet}  
\usepackage{courier}  
\usepackage[hyphens]{url}  
\usepackage{graphicx} 
\usepackage{natbib}  
\usepackage{caption} 
\usepackage{rotating}
\usepackage{tabularray}

\title{Overhearing LLM Agents: A Survey, Taxonomy, and Roadmap}

\author{
    Andrew Zhu
    and
    Chris Callison-Burch
}
\affiliations{
    University of Pennsylvania \\
    \texttt{\{andrz,ccb\}@seas.upenn.edu}
}

\usepackage{xcolor}
\usepackage{xifthen}
\newcommand\todo[1]{
    \textcolor{red}{
        \ifthenelse{\isempty{#1}}
        {TODO}
        {TODO: #1}
    }
}

\begin{document}

\maketitle

\begin{abstract}
    Imagine AI assistants that enhance conversations without interrupting them: quietly providing relevant information during a medical consultation, seamlessly preparing materials as teachers discuss lesson plans, or unobtrusively scheduling meetings as colleagues debate calendars.
    While modern conversational LLM agents directly assist human users with tasks through a chat interface, we study this alternative paradigm for interacting with LLM agents, which we call ``overhearing agents''.
    Rather than demanding the user's attention, overhearing agents continuously monitor ambient activity and intervene only when they can provide contextual assistance.
    In this paper, we present the first analysis of overhearing LLM agents as a distinct paradigm in human-AI interaction and establish a taxonomy of overhearing agent interactions and tasks grounded in a survey of works on prior LLM-powered agents and exploratory HCI studies.
    Based on this taxonomy, we create a list of best practices for researchers and developers building overhearing agent systems.
    Finally, we outline the remaining research gaps and reveal opportunities for future research in the overhearing paradigm.
\end{abstract}

\section{Introduction}
A teacher leads a classroom discussion about photosynthesis. As students ask questions, an overhearing agent silently recognizes knowledge gaps and queues up relevant diagrams to appear on the smart board at just the right moment. During a family dinner conversation about weekend plans, the agent notices consensus forming around a hiking trip and quietly prepares weather forecasts and trail recommendations for later review. In a medical consultation, as a patient describes symptoms to their doctor, the agent retrieves relevant case histories and recent research, displaying them unobtrusively on the physician's tablet. These scenarios illustrate the premise of overhearing agents: AI systems that enhance human activities without interrupting them, and assist without being asked.

There has been a recent uptick in the popularity of large language model (LLM)-powered AI agents: semi-autonomous systems that use multiple rounds of tool calling to answer complex queries or complete tasks delegated by a human user.
These agents are usually presented as chatbots, which the user converses with directly to give directions and receive results~\cite{li-2025-review}. We call this pattern of interaction, in which the agent is an active participant in a conversation, ``\textit{conversational} agents". However, directly conversing with an LLM may not be practical in some situations where a human would appreciate help from an AI agent.

\begin{figure}[t]
    \centering
    \includegraphics[width=0.95\columnwidth]{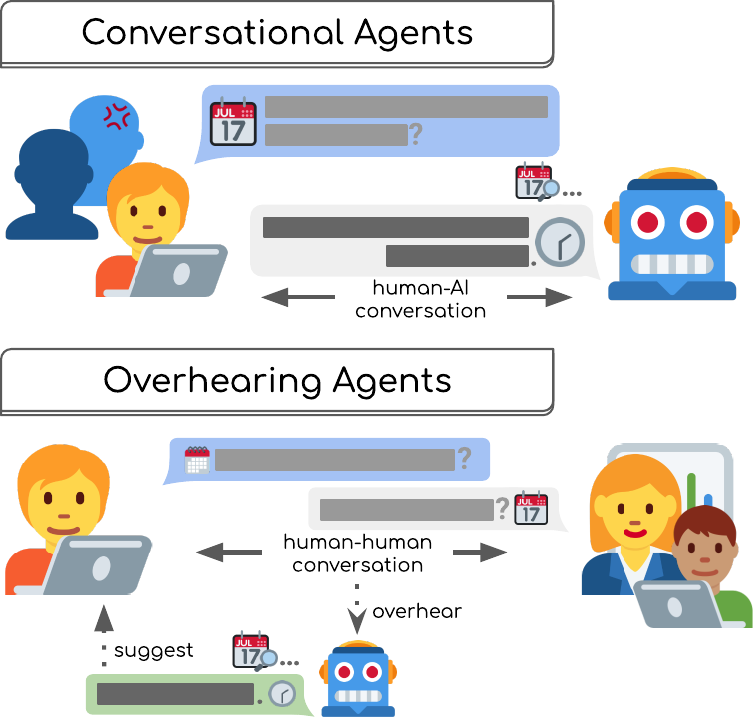}
    \caption{Conversational AI agents (top) are presented as a chatbot with which the user converses directly. We present \textit{overhearing} AI agents (bottom), which instead ``listen in'' on human-to-human conversations, and provide suggestions to the user without needing to participate in the conversation.}
    \label{fig:fig1}
\end{figure}

Thus in complement to conversational agents, we study \textit{overhearing} agents as an alternate paradigm for interacting with LLM agents. Rather than having human users directly converse with an LLM agent, the agent instead ``listens'' to a conversation among multiple human users and passively provides suggestions or takes actions in the background to assist (Figure~\ref{fig:fig1}).
To accomplish this, the overhearing agent must establish beliefs about the user's intentions, without the ability to ask them directly. This makes the task more difficult than that of a conversational agent~\cite{schober_understanding_1989, novick_mutual_1993}.

To realize this vision, we provide the first survey of overhearing LLM agents as an emerging paradigm for human-AI interaction. The contributions of this work are as follows:
\begin{itemize}
    \item We present a \textbf{survey and taxonomy} of overhearing agent systems grounded in prior conversational agent works and theoretical HCI studies.
    \item Based on our survey and taxonomy, we identify \textbf{design principles and interaction considerations} for building effective overhearing agents.
    \item Finally, we outline future \textbf{research directions and open challenges} in overhearing agent engineering, human-AI interaction, and software engineering.
\end{itemize}
We hope this survey can establish a foundation for future research and provide researchers with a roadmap for advancing this underexplored interaction paradigm.

\section{Related Work}

The concept of overhearing agents builds upon several established research areas while introducing unique challenges. We review relevant work in multiagent communication, autonomous agents, proactive systems, and existing AI assistance paradigms.

\paragraph{Overhearing in Multiagent Communication}
The idea of using overhearing agents as passive suggesters was initially introduced as a proposal to improve networked computer systems~\cite{busetta_extending_2001, aiello_ontological_2002, gutnik_towards_2004}. These early proposals involved an architecture in which service agents, carrying out some task, were supported by other agents, which gain knowledge about the goals of the service agents by ``listening'' to an abstract message stream between them. These overhearing agents are assumed to understand the task being carried out and can propose \textit{suggestions} to assist the service agents. We borrow the term ``overhearing agents'' from these works, using them as theoretical grounding in the AI agent domain, with human users acting as service agents and LLM agents acting as overhearing agents.

\paragraph{Ubiquitous Computing}
Ubiquitous computing envisions technology that provides ambient assistance while fading into the background of everyday life~\cite{weiser_computer_1999}. Home voice assistants like Amazon Alexa, Google Home, and Siri are often viewed as partial realizations of this vision, embedding computational capabilities into our living spaces.
More recently, the overhearing paradigm has been explored in the context of these voice assistants. While these assistants are usually activated by a wake word, exploratory studies have proposed that they could be used as overhearing agents~\cite{tabassum_investigating_2019}.

\paragraph{AI Copilots}
In the domains of writing and coding, AI ``copilots'' support their human user with passive suggestions. This domain is similar to that of overhearing agents, in that the AI agent ``overhears'' the user writing and makes suggestions in the form of proposed completions~\cite[\textit{inter alia}]{chen2021evaluatinglargelanguagemodels, yuan-2022-wordcraft, barke-2023-grounded}. However, the interaction with the copilot is done in a one-person setting, with the user writing a document by themselves. Thus, the copilot's observations do not include human conversational input.

\paragraph{Proactive Agents}
Proactive dialogue systems guide the user towards a system-defined objective during a conversation. These objectives range from discussing a target topic to accomplishing a user-defined task to eliciting specific information from the user~\cite{deng_survey_2023}. Recent works propose extending proactive dialogue systems to initiate a conversation with the user themselves based on an AI agent's observations of its environment~\cite{zargham_understanding_2022, lu_proactive_2024}. These types of proactive agents are similar to overhearing agents in that they predict when a user might need help working towards a certain objective based on environmental observations, but the interaction that follows takes the form of a conversation between the user and the assistant.



\paragraph{Autonomous LLM Agents}
With the ability to use tools and interact with the wider world, a popular topic in current work is LLM-powered ``agentic'' systems.
LLM agents use tool calling across multiple rounds of tool use in pursuit of a user-defined goal. Like traditional AI agents, these tools allow the model to perceive (in the form of retrieving information and knowledge) and interact with (e.g., clicking buttons or typing input) with the external world to accomplish the given goal.

Recently, autonomous conversational AI agents have gained popularity for a wide variety of use cases.
In the code-writing domain, tools like Cursor\footnote{\url{https://cursor.com/en}}, Claude Code \cite{claude_code_2025}, or MetaGPT \cite{hong2024metagpt} orchestrate AI agents to perform wide-reaching, multi-round edits of code on a repository-level scale. In everyday research and information retrieval, tools like Gemini\footnote{\url{https://gemini.google/overview/deep-research/}} or Claude Deep Research \cite{claude_research_2025}, AI2's ScholarQA \cite{singh-etal-2025-ai2}, and many others use a hierarchical multi-agent setup to aggregate a large number of sources to return an answer to a user's query.
These tools are all conversational agent systems which kick off a long-running process to generate a sequence of tool calls to write code, retrieve internet sources, or interface with public APIs (e.g. Semantic Scholar), with the exact task delegated to the system directly by the user.
These agents represent a pivot away from agents acting as an assistive tool for a human user, towards the AI agent performing tasks autonomously for the user given explicit guidance. However, we believe that using AI agents to assist users passively without explicit directions remains an underexplored area of research. This motivates us to explore the overhearing paradigm.

\begin{table*}[t]
\centering
\small
\begin{tblr}{
  cell{2}{1} = {r=10}{},
  cell{2}{2} = {r=4}{},
  cell{6}{2} = {r=3}{},
  cell{9}{2} = {r=3}{},
  cell{12}{1} = {r=6}{},
  cell{12}{2} = {r=2}{},
  cell{14}{2} = {r=2}{},
  cell{16}{2} = {r=2}{},
  vline{2-4} = {},
  hline{1,2,12,18} = {-}{0.12em},
  hline{3-5,7-8,10-11,13,15,17} = {3-4}{dotted},
  hline{6,9,14,16} = {2-4}{},
}
                                                        & \textbf{Dimension}      & \textbf{Value}    & \textbf{Example(s)} \\
\begin{sideways}\textbf{User Interaction}\end{sideways} & \textbf{Initiative}     & Always Active     & { Always-listening voice assistants \cite{tabassum_investigating_2019, zargham_understanding_2022} } \\
                                                        &                         & User-Initiated    & Wordcraft \cite{yuan-2022-wordcraft} \\
                                                        &                         & Post-Hoc Analysis & More to Meetings \cite{mcgregor_more_2017} \\
                                                        &                         & Rule-Based        & Apple Intelligence \cite{Apple_2025} \\
                                                        & \textbf{Input Modality} & Audio             & Bardo \cite{padovani_bardo_2017}, LlamaPIE \cite{chen2025llamapieproactiveinearconversation} \\
                                                        &                         & Text              & { Code ``Copilots'' \cite{barke-2023-grounded} } \\
                                                        &                         & Video             & EgoExo4D \cite{grauman_ego-exo4d_2024} \\
                                                        & \textbf{Interfaces}     & Web and Desktop   & { Proactive Search \cite{andolina_investigating_2018}, ProactiveAgent \cite{lu_proactive_2024} } \\
                                                        &                         & Wearable Devices  & LlamaPIE \cite{chen2025llamapieproactiveinearconversation}, Meta smart glasses \cite{meta2024rayban} \\
                                                        &                         & Smart Home        & { Home voice assistants  \cite{tabassum_investigating_2019, zargham_understanding_2022, hwang_rewriting_2023} } \\
\begin{sideways}\textbf{System Design}\end{sideways}    & \textbf{State}          & Read-Only         & Proactive Search \cite{andolina_investigating_2018} \\
                                                        &                         & Read-Write        & More to Meetings \cite{mcgregor_more_2017} \\
                                                        & \textbf{Timeliness}     & Real-Time         & Bardo \cite{padovani_bardo_2017} \\
                                                        &                         & Asynchronous      & ReDel \cite{zhu_redel_2024} \\
                                                        & \textbf{Interactivity}  & Foreground        & Most overhearing agents \\
                                                        &                         & Background        & AI agent memory \cite{park_generative_2023}, DocPrompting \cite{zhou_docprompting_2023}
\end{tblr}
\caption{An overview of our taxonomy of overhearing dimensions.}
\label{tab:taxonomy}
\end{table*}

\section{Dimensions of Overhearing}
Rather than having human users directly converse with an LLM agent, an overhearing agent instead ``listens'' to a conversation among multiple human users and passively provides suggestions or takes actions in the background to assist. ``Listening'' may directly refer to the agent processing conversational audio data from a recorded human conversation, or other modalities like text or video.
To accomplish the overhearing task, the overhearing agent must establish beliefs about the user's intentions without the ability to ask them directly, and determine if it has the appropriate tools to assist the user~\cite{schober_understanding_1989, novick_mutual_1993}.
There may be long stretches of time in which the user does not require any assistance from an overhearing agent, and times when the user requires sudden timely intervention from the agent.

We envision a system where the suggestions and actions that the agent can make/perform are defined through provided functions that the agent can choose to call. Unlike a conversational agent, the agent's outputs are never shown directly to the user. This is because as the user is occupied with another conversation, they would not have the time to inspect verbose conversational output from an AI agent. Instead, overhearing agents should use its language output for ``thinking''. This allows the agent to leverage inference-time compute to reason about the current conversation before deciding whether or not to call a tool, which has been shown to improve the  performance of AI agents~\cite{yao2023react}.


In order to provide a foundation for researchers to move towards such a system, we present a survey and derived taxonomy of human interaction with overhearing agents and the kinds of tasks performed by overhearing agents, grounded in existing conversational agent works and exploratory HCI studies. We divide this taxonomy into two main categories: how users interact with overhearing agents (§\ref{sec:iface-taxonomy}) and system design of overhearing agents (§\ref{sec:task-taxonomy}). We provide an overview of our taxonomy in Table \ref{tab:taxonomy}.

\section{User Interaction With Overhearing Agents}
\label{sec:iface-taxonomy}
How users interact with overhearing agents represents a primary consideration in their design and deployment. In this section, we examine three dimensions of user interaction: initiative (when agents become active), input modality (what types of information they process), and interfaces (how they communicate with users).

\subsection{Initiative}
\label{sec:initiative-taxonomy}
Initiative refers to the circumstances under which an overhearing agent becomes active and begins processing user interactions, and whether suggestions are proactively proposed by the overhearing agent or elicited by the user.



\paragraph{Always Active}
The most straightforward approach is for the AI agent to continuously monitor all conversations while on, and provide suggestions whenever the AI agent deems it helpful. While this ensures the agent never misses an opportunity to help, it raises significant privacy concerns and requires constant computational resources.
The agent must also implement relevance filtering to prevent suggestion fatigue.
This is particularly challenging because many modern AI agents are built with the assumption that they will only be invoked when the user requires their assistance. In contrast, overhearing agents may receive a large amount of input without needing to make a suggestion.

Most commonly, an always active overhearing agent is active during the full duration of an associated task it is specialized for, such as the course of a game it can assist with~\cite{padovani_bardo_2017, ferreira_computer-generated_2020, kelly_towards_2023}, writing code~\cite{chen2021evaluatinglargelanguagemodels, barke-2023-grounded}, a customer support chat~\cite{banerjee_system_2023}, or casual conversation~\cite{andolina_investigating_2018}. Additional studies propose environments in which an overhearing agent is truly always active, such as when it is embodied in a smart speaker~\cite{zargham_understanding_2022} or a helper application on a computer~\cite{lu_proactive_2024}.


\paragraph{User-Initiated}
The user-initiated approach instead requires the user to request assistance when they anticipate needing help. In an overhearing system, this must be done non-intrusively, for example with a button press on a smartwatch.
While this works well for discrete tasks, it may miss important context if activated after the user desired assistance.
Given audio or video inputs, an overhearing agent might need to implement a rolling buffer of recent audio/video that it can use to ground its suggestions when activated.
Given text inputs, an overhearing agent can refer to previously written text without a need for an additional buffer. An example of a user-initiated overhearing agent is Wordcraft~\cite{yuan-2022-wordcraft}, which proposes a continuation to user-written text when requested with a button press.

\citet{zhu_first_2025} find that a user-initiated approach results in one-fifth the activations as an always-active approach. This suggests that asking the user to initiate suggestions may effectively reduce suggestion fatigue. However, this may be a trade-off between precision and recall, as there may be instances of aid where the user may not be able to pinpoint an exact time to activate an overhearing agent.

\paragraph{Post-Hoc Analysis}
\label{sec:post-hoc-analysis}
Post-hoc systems process conversations after they conclude rather than in real-time. Users explicitly upload recordings for analysis, which offers complete context and allows more sophisticated processing without real-time constraints. However, this sacrifices immediate in-conversation assistance, making it unsuitable for time-sensitive support.
This approach is commonly seen in commercial ``AI note-taking'' products which send meeting participants a list of identified action items and a summary of a meeting post-hoc~\cite{mcgregor_more_2017}. Other ``life-logging'' products like Bee\footnote{\url{https://www.bee.computer/}} and Compass AI\footnote{\url{https://shop.compasswearable.com/}} provide a summary of your day and suggested to-dos at the end of the day based on transcripts collected throughout the day.

\paragraph{Rule-Based}
Rule-based activation uses environmental or behavioral signals to automatically engage the overhearing agent. These systems might activate when entering specific locations like conference rooms, during calendared meetings, or when particular applications are open. For example, Apple Intelligence proactively highlights important notifications when a rule-activated focus mode is active on an Apple device~\cite{Apple_2025}.


\subsection{Input Modality}
Overhearing agents can operate across different input modalities, each offering distinct types of information and affordances. While these modalities are not mutually exclusive---many systems may combine multiple inputs for richer understanding---each has unique characteristics that shape how overhearing agents function.

\paragraph{Audio}
Audio input is the most natural modality for overhearing agents, as it directly captures human conversation. Beyond the explicit content of speech, audio provides rich implicit cues that help agents determine when assistance is needed. These include prosodic features like pauses, hesitations, or changes in tone that might indicate confusion or uncertainty. Audio agents may implicitly diarize speech to distinguish between participants, and environmental sounds can provide context about the setting and activity~\cite{zhu_first_2025}.

Most prior work does not use audio input directly, instead transcribing and optionally diarizing speech before using an LLM to perform the downstream task~\cite{padovani_bardo_2017, mcgregor_more_2017, andolina_investigating_2018, ferreira_computer-generated_2020, kelly_towards_2023, chen2025llamapieproactiveinearconversation}. This is because until recently, there have been few multimodal language models capable of processing audio input directly~\cite{openai2024gpt4ocard, fixie2025ultravox, xu2025qwen25omnitechnicalreport, microsoft2025phi4minitechnicalreportcompact, google_gemini_2025}.


\paragraph{Text}
Rather than ``listening'' to a conversation, text-based overhearing agents monitor text-based chat rooms~\cite{banerjee_system_2023}, documents a user is writing~\cite{yuan-2022-wordcraft, barke-2023-grounded}, or other structured textual/code inputs (e.g., an event log; \citealt{lu_proactive_2024}). Unlike conversational overhearing, text agents more frequently operate in single-user contexts, ``overhearing'' the user's interaction with a document or interface.

While text modality lacks the implicit cues available in audio, it offers perfect transcription accuracy and implicitly includes the context the user is working in (e.g., the code repository or surrounding text document) without having to record a buffer of speech. This can make the overhearing agent's intent prediction task easier by grounding the task the user is working on in this included surrounding context.

\paragraph{Video}
Video input enables agents to perceive non-verbal communication, spatial context, and physical activities. For example, an overhearing agent might help a user who is cooking by identifying a completed recipe step, automatically setting timers, and displaying the next step~\cite{hwang_rewriting_2023}. Emerging datasets like Ego-Exo4D~\cite{grauman_ego-exo4d_2024} lay the groundwork for video-based overhearing agents by providing rich multimodal data of human activities from multiple perspectives with detailed expert annotations.


\subsection{Interfaces}

We examine several interface categories through which overhearing agents can surface suggestions to users. Each interface type offers different affordances to the user.
An overhearing agent might use a combination of these interfaces to effectively manage user input and suggestion output. For example, the system might take audio input from a wearable device and process it on another device with greater compute capability (or the cloud) before sending any relevant suggestions as notifications on the user's smartphone.

\paragraph{Web and Desktop Interfaces} can be used to display detailed suggestions and support complex interactions. These interfaces excel at presenting multiple suggestions simultaneously, showing full context, and enabling sophisticated editing workflows. Such interfaces range from a dedicated webpage/application for the overhearing agent~\cite{andolina_investigating_2018, yuan-2022-wordcraft, kelly_towards_2023} to integrations with existing applications~\cite{barke-2023-grounded, lu_proactive_2024} or a screen overlay.
For instance, a meeting assistant could display extracted action items, referenced documents, and suggested calendar events in a sidebar interface. The persistent nature of screen-based interfaces also allows users to queue suggestions for later review.

\paragraph{Wearable Devices} Smartwatches can deliver haptic notifications with glanceable text summaries, allowing users to privately review suggestions without disrupting conversation flow. Audio wearables (e.g., earbuds) could be used to provide short reminders when a user hesitates~\cite{chen2025llamapieproactiveinearconversation}. Augmented reality smart glasses could overlay relevant information in the user's perceptual space, like showing the next step of a recipe the user is cooking~\cite{hwang_rewriting_2023} or translating conversations in real time~\cite{meta2024rayban}.

\paragraph{Smart Home Devices} Though currently limited to wake-word activation, smart home devices have been identified as a promising platform for the overhearing paradigm~\cite{tabassum_investigating_2019, zargham_understanding_2022, hwang_rewriting_2023}. These devices are already deployed in millions of homes, and include the necessary hardware for audio recording, audio playback, and internet connectivity.


\section{Overhearing System Architecture}
\label{sec:task-taxonomy}
Beyond user interaction, the system architecture defines how overhearing agents process and execute tasks.
Overhearing agent tasks do not differ fundamentally from conversational agent tasks. Thus, we base our taxonomy of overhearing task dimensions grounded in existing conversational agent implementations.

\subsection{State}
The state dimension refers to whether an overhearing task modifies the user's environment.

\paragraph{Read-Only}
Read-only tasks do not modify any external resource or affect any future tasks. Most commonly, stateless tasks are information retrieval or calculation tasks.
Simple examples include providing additional information about entities or questions mentioned in conversation~\cite{andolina_investigating_2018, chen2025llamapieproactiveinearconversation, zhu_first_2025} or retrieving the user's calendar during scheduling discussions~\cite{lu_proactive_2024}.


\paragraph{Read-Write}
In contrast, read-write or stateful tasks require an AI agent to be aware of a long-term state and how its actions will affect that state (a ``world model''; \citealp{hao_reasoning_2023}). Completing a read-write task requires making modifications to the user's environment that impact future tasks.
An example of a read-write overhearing task would be the agent identifying when a conversation has reached a consensus on when to meet, and adding that time to the user's calendar~\cite{mcgregor_more_2017}. This modifies the user's environment (the calendar) and affects future tasks (as the calendar will now show the scheduled time as busy). Other examples include playing music to match the current mood of a game~\cite{padovani_bardo_2017, ferreira_computer-generated_2020}, or calling for aid if an emergency situation is detected~\cite{zargham_understanding_2022}.

\subsection{Timeliness}
Timeliness refers to whether an overhearing task must be completed in real time or can be completed at any time after the recognition of support.


\paragraph{Real-Time}

Real-time overhearing tasks must be completed as soon as the user requires assistance, or their utility significantly diminishes.
Examples of synchronous overhearing tasks include displaying the next step in a recipe when a user has finished the current step~\cite{hwang_rewriting_2023}, playing a sound effect in a tabletop game~\cite{padovani_bardo_2017}, or running a search for a recently mentioned topic~\cite{andolina_investigating_2018}.

\paragraph{Asynchronous}
In contrast, an asynchronous task does not necessarily need to be completed in real time---it could spawn a long-running task to generate an artifact to be reviewed by the user in the future or take a non-critical action that could be approved at any time.
Asynchronous tasks are superficially similar to the \textit{Post-Hoc Analysis} dimension introduced in Section~\ref{sec:post-hoc-analysis}. However, whereas a post-hoc system can only make suggestions when the user uploads a conversation for analysis, asynchronous tasks can be launched during the course of a conversation for later review and are compatible with any Initiative dimension.

Examples of asynchronous overhearing tasks include suggesting to add an event to a calendar during a meeting (which the user can edit the details of and accept or reject after the meeting; \citealp{mcgregor_more_2017}), writing to an agent's long-term memory~\cite{park_generative_2023}, or launching a long-running multi-agent system process to generate a report for later review~\cite{zhu_redel_2024}.

\subsection{Interactivity}

The term "interactivity" refers to whether an LLM uses a tool call to send information to the user or to update its beliefs silently in the background.

\paragraph{Foreground}
Foreground tasks are when the agent is interacting with the user by making a suggestion. This is the primary method by which overhearing agents interact with their users.

\paragraph{Background}
In contrast, a background task refers to when an overhearing agent performs an agentic task (i.e., uses multiple tool calls and interacts with the environment) to update its internal world model only, without making any suggestion to the user. The overhearing agent might choose to do this proactively to improve its ability to predict the user's intent by retrieving additional context. For example, an overhearing agent might use a tool call to retrieve logs of previous interactions to ground a currently ongoing conversation in~\cite{park_generative_2023}. Similarly, a coding copilot might use a tool call to retrieve documentation or internal libraries included in a code repository that is not in its context window~\cite{zhou_docprompting_2023}.

\section{Developing Overhearing Agents}
Building effective overhearing agents requires addressing unique technical and design challenges beyond those faced by traditional conversational AI. In this section, we examine key considerations for future researchers and developers and make suggestions for best practices for user privacy, maximum utility, and developer experience.

\subsection{Privacy \& Security}

In this paper, we envision how LLM-powered AI agents can passively listen in on human conversations and make relevant suggestions.
In order to accomplish the overhearing agent task, such agents must record and process a large amount of ambient data, which may include audio or video from the user's private spaces (§\ref{sec:iface-taxonomy}).
\citet{tabassum_investigating_2019} showed that less than half of a surveyed population would feel comfortable allowing such recording from an ``always-listening'' device. In public, such a device may record passers-by without their knowledge or consent. Recorded multimodal inputs may also be subject to legal discovery in case of an investigation, which could be used for widespread surveillance of a population by an authoritarian state \cite{peterson_risky_2023, scammell_lawsuit_2025}.

Audio or other multimodal recordings which are stored for an overhearing agent must also take into account data security considerations, as the recorded inputs may contain PII or other private information.
In certain settings (e.g., corporate or medical), it may not be desirable or legal to disclose recorded data to a third party for AI processing.

With these considerations in mind, we make the following suggestions. Stored inputs should \textbf{redact PII} where possible (for example, using a tool such as Microsoft Presidio \cite{microsoft_presidio_2024}), and any locally-recorded data should be \textbf{at-rest encrypted}. Users should have the option for \textbf{on-device processing} with small language models rather than API-hosted large language models. Finally, systems should implement \textbf{clear consent mechanisms} that inform all parties when recording.

\subsection{User Interaction Considerations}
\label{sec:interaction-considerations}
In Section~\ref{sec:iface-taxonomy} we presented a taxonomy of the user interfaces that an overhearing agent could use to receive input from a user and surface suggestions to a user.
Recall that an overhearing agent does not directly converse with the user---it only makes suggestions that the user can choose to accept or reject, while the user is engaged in an activity that demands their primary focus.

As such, we propose the following design principles for user interaction to make overhearing agents present maximum utility for their user without being overbearing.

First, a suggestion from an overhearing agent should be \textbf{verifiable at a glance}. If relevant, more detailed information could be displayed to the user upon interaction (e.g., tapping a notification, clicking on a link).
Second, suggestions must be \textbf{dismissible without friction}. Since overhearing agents make proactive suggestions without explicit user request, false positives are inevitable.
Third, overhearing agents should support \textbf{reversible actions}. When users accept a suggestion that modifies state (such as adding a calendar event), they should be able to easily undo that action if they later determine it was incorrect or unnecessary.
Fourth, suggestions that create content or take an action should be \textbf{editable}. This allows an overhearing agent that correctly identifies user intent but imperfectly executes to provide partial value to the user. For example, a user could correct a meeting's participant list or modify transcribed text.
Finally, overhearing agents should manage multiple suggestions with an \textbf{intelligent queuing system}. The interface should prioritize time-sensitive suggestions while allowing users to pin important items for later review and automatically expiring stale suggestions that are no longer relevant. For example, providing the user with references from a document the user is actively speaking about is more time-sensitive than adding a calendar event.

\subsection{Building Tools for Overhearing Tasks}
The effectiveness of overhearing agents depends on their ability to access and manipulate external information through tools.
In Section~\ref{sec:task-taxonomy}, we presented a taxonomy of the types of tasks that an overhearing AI agent could perform.
This taxonomy can guide the types of tools provided to an overhearing agent. An overhearing agent should use the tool definitions given to it to determine what kinds of tasks it is capable of and when it would be appropriate to use these tools to provide suggestions to the user.


Given these principles and the system design dimensions outlines above, we make the following recommendations. A tool interface should allows developers and researchers to define tools in \textbf{plain Python code}, which allows for developers to use familiar libraries and easily integrate an AI agent with external APIs or local data. A tool interface should implement a \textbf{modular design} to allow for subsets of tools to be added or removed from an LLM's context window dynamically depending on the downstream use case. Tool interfaces should be \textbf{LLM-agnostic}, allowing a single set of tools to be reused regardless of the underlying model. Finally, tool interfaces should be built with an \textbf{asynchronous-first} design, allowing for both real-time and asynchronous tasks to be implemented using native Python programming models.
Previous works on LLM tool calling such as Kani~\cite{zhu-etal-2023-kani} and the Model Context Protocol\footnote{\url{https://modelcontextprotocol.io/}} (MCP) present first steps towards these goals.


\section{Research Challenges and Future Directions}

In this survey, we categorized the various dimensions of overhearing agents and examined best practices for building such systems. However, significant research challenges remain. We outline five key areas for further research:


\paragraph{Challenge 1. \textit{How can overhearing agents predict optimal intervention points from a continuous overheard conversation?}}
During a conversation which an overhearing agent is listening to, the overhearing agent is fed with a continuous stream of input. To assist the user, the agent must predict from this input stream when to intervene with a suggestion in a timely manner. There are two main approaches to tackle this challenge: explicit segmentation of the continuous input into discrete turns using semantic VAD models~\cite{skantze_turn-taking_2021, kim_voice_2018, shi_semantic_2023}, and parallel processing of the continuous input stream and an output stream~\cite{defossez_moshi_2024}.

Based on explicit segmentation of continuous input, further work could build a multimodal semantic model with the objective to predict possible intervention points directly from a continuous input stream. Based on continuous parallel processing, future work could investigate how to integrate agentic tool calling with a full-duplex system.


\paragraph{Challenge 2. \textit{How do we evaluate the helpfulness of an overhearing agent?}}
To assist its human user, an overhearing agent must predict the user's intentions from the user's behaviour and previous interactions. This is inherently a difficult task, and it is likely that overhearing agents' suggestions will be imperfect: they might miss intervention points, provide spurious suggestions, or suggest content that is only partially correct (§\ref{sec:interaction-considerations}). User studies have shown that even imperfect suggestions are helpful to users in low-stakes brainstorming and passive assistance scenarios~\cite{kocielnik_will_2019, weisz_perfection_2021, yuan-2022-wordcraft, barke-2023-grounded}. 

In the overhearing setting, users may be focused on a task that demands their focus and may not appreciate the additional cognitive load of verifying spurious suggestions from an overhearing agent. Thus, it is important to create metrics to evaluate the utility that an overhearing agent provides.


\paragraph{Challenge 3. \textit{How can we optimize multimodal throughput for real-time processing?}}
In order to perform the real-time overhearing task, a multimodal language model must be capable of processing a continuous stream of audio and/or video at at least real-time speed. Current multimodal language models accomplish this by encoding a fixed duration of multimodal input into a fixed number of tokens, which are processed by a Transformer-based model to predict a sequence of output tokens. While optimizing the Transformer model to reduce prefill and generation time is widely researched~\cite[\textit{inter alia}]{dettmers_llmint8_2022, kwon_efficient_2023, devvrit_matformer_2024}, optimizing the representation of multimodal input remains an open challenge.

In overhearing systems, the information density of a continuous input stream may not be constant, as there may be periods of silence or stillness in audio or video. Thus, future work could develop a variable rate tokenization scheme that varies with the information density of a continuous input. For example, fast speech in a noisy environment might encode each second of audio as many tokens, while a long span of silence might only use a few tokens. Such a tokenization scheme could improve the throughput of a multimodal system while simultaneously reducing the cost associated with processing long context sequences.


\paragraph{Challenge 4. \textit{How do we design software libraries to support the research and development of overhearing agents?}}
As a developer building an overhearing agent, there are additional dimensions to consider beyond the interfaces exposed by most modern AI agent software libraries. Most modern libraries are built with a text-first, desktop-first, synchronous design in mind and do not support many of the dimensions of overhearing agents outlined in this paper.
A software library built for overhearing agents should support native audio and video I/O, integration with mobile devices and their device-specific affordances (e.g., mobile notifications and haptics), and asynchronous programming. These programming tools should be provided to the developer as a high-level interface to reduce the overhead of device-specific integration, allowing for faster prototyping and iteration of overhearing tasks and prompts instead.

\paragraph{Challenge 5. \textit{How can overhearing agents negotiate consent in multi-party settings?}}
Although overhearing agents primarily serve a single user, they must process conversations involving multiple participants who may not consent to being recorded or analyzed. Today's approach primarily falls into one of two categories: ``least permissive'' systems which only allow recording if all participants consent, or systems that record by default and assume non-consenting participants will not interact with the system. Future research could focus on selective processing privacy approaches in the middle~\cite{barhm_negotiating_2011}, to make overhearing agents maximally useful while allowing individuals to opt out of the recording or processing of their individual data.

\section{Conclusion}
In this paper, we presented a survey of overhearing agents: AI systems that passively monitor human conversations to provide contextual assistance without direct participation. We introduced a taxonomy encompassing key dimensions of overhearing systems: initiative patterns, input modalities, and task architectures.
Based on this taxonomy, we presented a set of best practices for designing secure and privacy-centric overhearing systems, interface considerations for surfacing suggestions to users, and development principles for building tools for overhearing LLM agents.
Finally, we outlined five open research challenges for these agents to maximize their potential while respecting human values and privacy.

Overhearing agents face unique challenges compared to conversational agents. They must infer user intent without explicit delegation, predict optimal intervention points from continuous input streams, and balance helpfulness against interruption.
As multimodal language models advance, overhearing agents show promise across diverse applications from supporting educators in classrooms to game masters in tabletop games and clinicians during procedures. We hope this paper provides both a foundation for researchers and a roadmap for addressing these open challenges, ultimately enabling more natural AI assistance that enhances human activities.


\bibliography{custom}


\newpage 
\appendix

\section{Input Segmentation for Overhearing}
\label{sec:app:segmentation}
In the text modality, explicit input segmentation is usually accomplished by design as the user sends discrete ``messages'' to the LLM.
Recent audio models extend this approach and use naive (e.g., tail-silence based) or semantic VAD models~\cite{kim_voice_2018, shi_semantic_2023} to segment a continuous audio stream into discrete ``messages''. After receiving a user message, the LLM then generates a response before awaiting the next user message~\cite{skantze_turn-taking_2021}.
Although advertised as full-duplex audio-language models, most such systems are in fact half-duplex systems with low enough latency to maintain the illusion that they are processing input while the user is still speaking. Thus, we call these systems ``pseudo-full-duplex''.

A more naive method is simply to send input to the language model in fixed-duration chunks, under the assumption that if a current chunk does not contain all necessary context for predicting a suggestion, the model will simply wait for the next chunk.
Explicit segmentation in this manner makes the challenge of deciding when to make a suggestion and calling tools easier, as overhearing systems can leverage the turn-based conversational fine-tuning of instruction-tuned language models to predict whether to intervene at discrete entrypoints. However, relying on explicit turn-based segmentation may introduce latency to an overhearing system and may not generalize to additional modalities such as video.

Truly full-duplex systems have been built for simultaneous processing of a continuous audio input stream and output language tokens, but these systems are built for conversation and struggle with knowledge or agentic tasks~\cite{defossez_moshi_2024}.





\section{Human Creativity \& Misuse}

As of writing, there has been a recent controversy involving a student at Columbia University using a passive multimodal AI tool to ``cheat'' interviews involving programming skill assessment at major US corporations \cite{elkins_cheating_2025, yang_kicked_2025}. This tool, created by the student, reportedly uses audio and video recordings of ongoing technical interviews to provide answers to interview questions in real time. While this is undoubtedly an overhearing agent task as defined in this work, such a use case differs dramatically from the tasks studied in this paper. We examine the use of passive AI agents to \textit{aid} humans in non-creative background tasks which would otherwise detract from their main creative goal, instead allowing them to dedicate more of their focus towards that goal. In contrast, the tool under scrutiny is designed to \textit{replace} a human in matters of intellectual ideation, using the human solely as a communication channel. It has been shown, particularly in creative domains, that the use of LLMs for ideation results in less interesting, more homogeneous creations~\cite{see_massively_2019, ippolito_creative_2022, anderson_homogenization_2024}. Although overhearing agents can be used to ``cheat'' on day-to-day tasks expected of a human, this is both ethically and creatively dubious. We encourage all deployments of overhearing agents to focus on tasks that aid, rather than replace, human creativity.

\section*{Acknowledgments}

Thank you to Princess Sampson and Evan Osgood for insightful conversations about the overhearing paradigm, help with HCI principles, and feedback on various drafts of the paper.

The research mentioned in this report is supported in part by the Office of the Director of National Intelligence (ODNI), Intelligence Advanced Research Projects Activity (IARPA), via the HIATUS Program contract \#2022-22072200005 and Defense Advanced Research Projects Agency’s (DARPA) SciFy program (Agreement No. HR00112520300).
This material is based upon work supported by the National Science Foundation Graduate Research Fellowship, under Grant No. DGE-2236662. The views and conclusions contained herein are those of the authors and should not be interpreted as necessarily representing the official policies or views, either expressed or implied, of ODNI, IARPA, DARPA, the NSF, or the U.S.\ Government. The U.S.\ Government is authorized to reproduce and distribute reprints for governmental purposes notwithstanding any copyright annotation therein.

\end{document}